\documentclass{article}

\usepackage{arxiv}

\usepackage[utf8]{inputenc} 
\usepackage[T1]{fontenc}    
\usepackage{hyperref}       
\usepackage{url}            
\usepackage{booktabs}       
\usepackage{amsfonts}       
\usepackage{nicefrac}       
\usepackage{microtype}      
\usepackage{lipsum}		
\usepackage{graphicx}
\usepackage{natbib}
\usepackage{doi}
\usepackage{microtype} 
\usepackage{fancyhdr}
\usepackage{graphicx} 

\usepackage{subfigure} 

\usepackage{booktabs} 
\usepackage{hyperref}

\title{DeepCompress: Efficient Point Cloud Geometry Compression}

\author{ Ryan Killea*\thanks{* Denotes corresponding author. We would like to recognize Per Karlsson, Jacob Str\"{o}m, \L{}ukasz Litwic, and Volodya Grancharov for their assistance and feedback. Replication code is available \href{https://github.com/pmclSF/DeepCompress}{here}. $\copyright$ Ericsson Inc 2021.} \\
	Worcester Polytechnic Institute\\
	Worcester, MA  \\
	\texttt{rbkillea@wpi.edu} \\
	\And
	Yun Li \\
	Ericsson Research\\
	Lund, Sweden \\
	\texttt{yun.y.li@ericsson.com} \\
	 \AND
	 Saeed Bastani \\
	 Ericsson Research \\
	 Lund, Sweden \\
	 \texttt{saeed.bastani@ericsson.com} \\
	 \And
	 Paul McLachlan \\
	 Ericsson Research \\
	 San Francisco, CA \\
	 \texttt{paul.mclachlan@ericsson.com} \\
}

\renewcommand{\shorttitle}{DeepCompress}

\hypersetup{
pdftitle={DeepCompress},
pdfsubject={q-bio.NC, q-bio.QM},
pdfauthor={Ryan Killea, Yun Y. Li, Saeed Bastani, Paul McLachlan},
pdfkeywords={Machine Learning, Point Cloud, Deep Learning, Compression},
}

\begin{document}
\maketitle

\begin{abstract}
Point clouds are a basic data type that is increasingly of interest as 3D content becomes more ubiquitous. Applications using point clouds include virtual, augmented, and mixed reality and autonomous driving. We propose a more efficient deep learning-based encoder architecture for point clouds compression that incorporates principles from established 3D object detection and image compression architectures. Through an ablation study, we show that incorporating the learned activation function from Computational Efficient Neural Image Compression (CENIC) and designing more parameter-efficient convolutional blocks yields dramatic gains in efficiency and performance. Our proposed architecture incorporates Generalized Divisive Normalization (GDN) activations and propose a spatially separable InceptionV4-inspired block. We then evaluate rate-distortion curves on the standard JPEG Pleno 8i Voxelized Full Bodies (8iVFB) dataset to evaluate our model’s performance. Our proposed modifications outperform the baseline approaches by a small margin in terms of Bjontegard delta (BD) rate and PSNR values, yet reduces necessary encoder convolution operations by 8\% and reduces total encoder parameters by 20\%. Our proposed architecture, when considered on its own, has a small penalty of 0.02\% in Chamfer’s Distance (D1) and 0.32\% increased bit rate in Point to Plane Distance (D2) for the same peak signal-to-noise ratio (PSNR). 
\end{abstract}


\section{Introduction} 

The importance of three-dimensional point cloud data for consumer and industrial applications is growing rapidly \citep{schwarz2018emerging}. Because point clouds are unstructured, they are distinct from other methods of representing three-dimensional geometry such as meshes or signed distance fields and digitize and represent any non-manifold structure. Many commercially available consumer devices have embedded 3D cameras, while fully and semi-autonomous vehicles incorporate LiDAR scans for environmental understanding and object tracking, such as simultaneous localization and mapping (SLAM) algorithms \citep{guo2020deep}. Augmented and virtual reality headsets also rely upon three-dimensional data to correctly place overlays and interactive content. These cameras and sensors generate three-dimensional point clouds, which are unstructured sets of coordinates in three dimensions. Even though they are unstructured, point cloud files are still extremely large.  

Many of the above use cases occur in compute- or battery-constrained environments, such as mobile phones or wearable headsets. This observation motivates our research on encoder efficiency for point cloud data. One might assume that a natural step would be to follow recent advances in deep learning image encoding and replace 2D convolution with 3D convolution and capture auto correlation in point cloud structure.  Running these models is extremely computationally demanding, especially on battery-powered devices such as mobile phones. In this paper, we show that an alternative network architecture can reduce parameter size and computation complexity with minimal impact on compression efficiency.

This paper capitalizes on the state-of-the-art in deep learning architectures for point cloud compression and aims to enhance architecture efficiency while preserving accuracy. We use deep learning to reduce the operational complexity for the compression of point cloud data. Deep learning successfully learns highly nonlinear functions that efficiently represent three-dimensional data, such as point clouds. Our model, which we refer to as DeepCompress, reduces the total number of operations by 8\% and total model parameters by 20\%, with negligible impact on compression efficiency. Using a standard test of efficiency — the Bjontegard delta (BD) rate — we only find a penalty of 0.02\% D1 and 0.32\% D2 increased bitrate for the same peak signal-to-noise-radio (PSNR).\footnote{We calculate this using the JVET spreadsheet which employs a cubic approximation to estimate the BD distance between curves.}  This result supports the use of point clouds in recourse- constrained environments and suggests additional research opportunities.   

\section{Prior work} 

Image compression with deep learning has been gaining significant advancements in recent years. Given the broad use of image classification in academia and industry, academic research on deep neural networks (DNNs) focuses heavily on improving image classification accuracy and efficiency. Yet, many emerging use cases — such as augmented and virtual reality and autonomous vehicles — are predicated on image classification and environmental understanding in extremely resource constrained compute environments. In addition to efficiency, many of these use cases also depend upon short model run times. Given these requirements, there is surprisingly little research on compressing an increasingly common data structure: point clouds. 

Traditional leaned 2D compression methods are not well adapted for point cloud data. This is due to substantial differences in data format between point clouds and image data, such as three-dimensionality, sparsity, and strict 0-1 occupancy. These differences between image data and point clouds motivates prior research on learned 3D representation, such as ResNets \citep{he2015deep} and Inception \citep{szegedy2014going, szegedy2016inceptionv4} through \citet{brock2016generative}. In the following subsection, we briefly discuss prior work on learned image compression and its applications to point cloud compression.

\subsection{2D Image Compression} 

Unlike 2D image or video data, point cloud data is highly sparse, has strict 0-1 occupancy, and is stored in three dimensions. Despite these differences in data, one can take architectural inspiration from 2D image compression techniques. Deep 2D image compression techniques use DNN transforms akin to 2D image classification. These models attempt to learn lossy image compression transforms in order to maximize peak signal-to-noise ratio (PSNR) or an image quality metric while minimizing the code’s entropy.  

\citet{balle2016endtoend} propose an end-to-end approach that uses a learned entropy model to jointly optimizes rate and distortion terms. They replace the non-differentiable transformation of quantization with uniform noise. \citet{balle2018variational} build on this work and proposes a learned scale hyperprior. These works are further extended by a mean and scale prior and with a context model, which have been shown to outperform the H.265 intra/BPG image coding scheme.  

Even though the bulk of previous 2D image compression research focuses on optimizing rate-distortion performance, recent work addresses the challenges of computational complexity and run time in deep learning architectures for image compression. \citet{johnston2019computationally} introduces Computationally Efficient Neural Image Compression (CENIC), which suggests a method to automatically optimize network architecture. They consider the optimization of GDN activation function and the use of regularization techniques. The learned compression techniques used in the works that we base our work off of use multiple learned image compression techniques, but do not use GDN because it was proposed nearly simultaneously. While this is an advancement on the state of the art, it also highlights the relative absence in the literature of DNN-based 2D image compression methods that optimize for efficiency.  

\subsection{3D Point Cloud Geometry Compression} 

3D point cloud compression methods fall into two categories: traditional algorithms and learned compression methods. Traditional algorithms for point cloud compression apply primitives from traditional video compression to 2D projections of the geometry (V-PCC) \citep{schwarz2018emerging}. V-PCC functions by grouping normals calculated for each point to find well-chosen sets of projections to minimize the overlapping of points after projection and optimized the smoothness of projected image for coding with a traditional video encoder such as H.265.Another approach directly compresses the 3D geometry using the autocorrelation in the octree structure (G-PCC Octree) \citep{schnabel2006octree}. The correlation can be exploited by using a wavelet transform or a graph Fourier transform. A third approach groups nearby points into triangles (G-PCC Trisoup) in regions where the point cloud is dense.   

Learned compression attempts to learn efficient compression codecs from 3D datasets with surface characteristics similar to point clouds. As it flexibly represents highly non-linear functions, deep learning is especially well-suited to point cloud compression (PCC). An earlier attempt to extend the learned image compression to the point cloud domain was by \citet{quach2019learning}, this suite of techniques is commonly referred to as learned PCC. The standard learned PCC methodology trains 3D convolutional neural network autoencoders (3D CNN AE) with a learned prior. Building on this base architecture, new research improves codec efficiency to reduce file size \citep{tang2020deep}, increase compression rates \citep{biswas2020muscle}, and reduce model parameters \citep{guarda2020deep}.  

The standard learned 3D CNN AE methodology uses a voxelized $[0,1]$ occupancy map representing a portion of the point cloud at a specific resolution as an input. The octree localizing each occupancy map within the full point cloud is then losslessly transmitted at the start or end of each point cloud. This transmission also includes other pieces of supporting information, such as scale within the broader environment and metadata.  

We consider the baseline PCC architecture from \citet{quach2020improved}, which incorporates several improvements upon \citet{quach2019learning}. These improvements include sequential training, focal loss, and more efficient architectural choices including residual blocks and progressively increasing channels as the resolution decreases. Their approach is distinct from other prior work focusing on graph-based representations of point clouds \citep{de2018graph}, multiscale approaches to compression \citep{wang2020multiscale}; lossless compression \citep{nguyen2020learning}, and vector-quantized approaches \citep{caccia2020online}. While these learned PCC approaches considerably improve rate distortion compared to traditional algorithms, these improvements come at the cost of slower run times and decreased computational efficiency.  

\begin{figure*}[ht] 

    \vskip 0.2in 

    \begin{center} 

    \centerline{\includegraphics[width=0.95\textwidth]{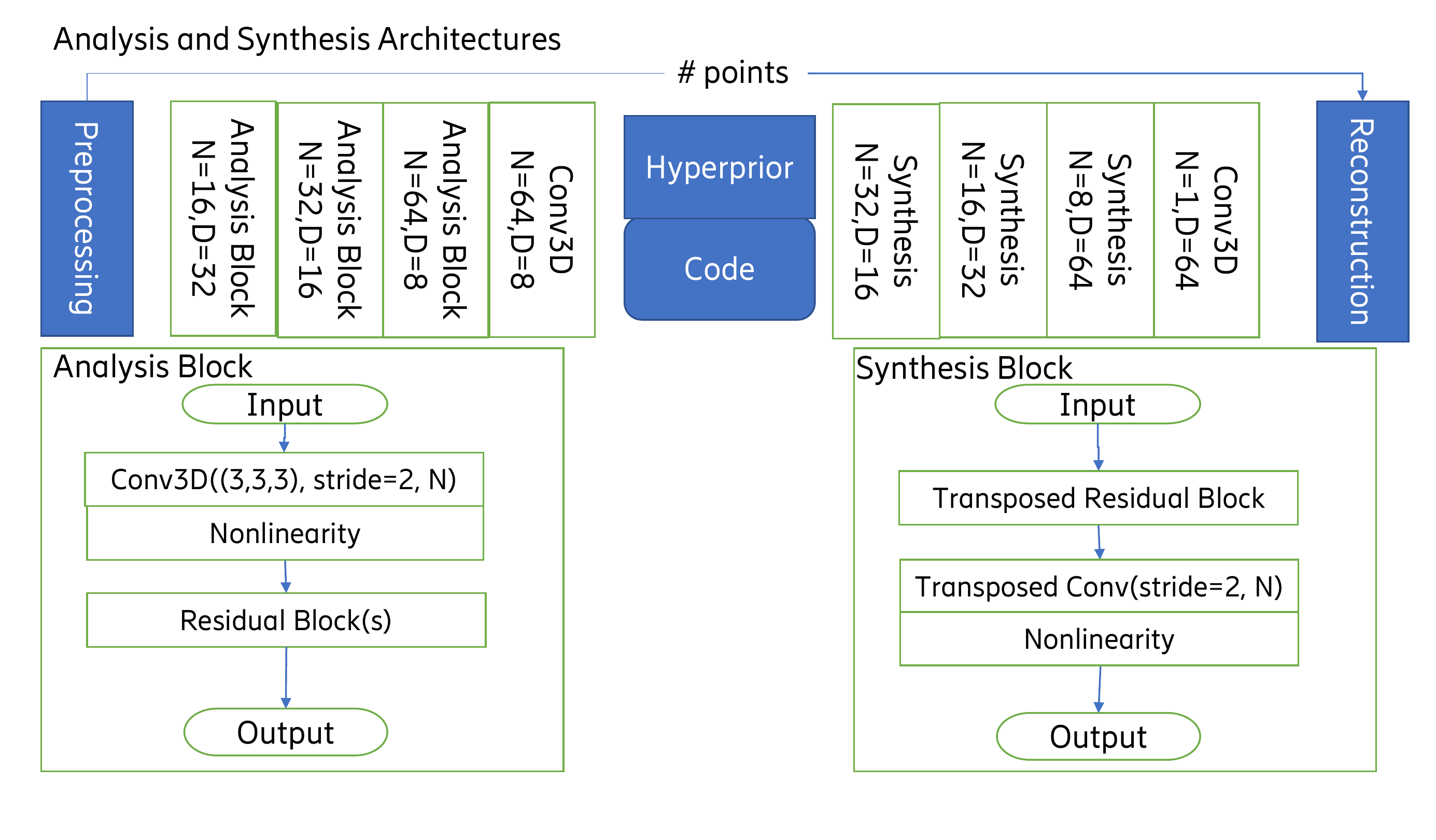}} 

    \caption{The baseline autoencoder network architecture we inherit. Because we are primarily concerned with improving our architecture’s efficiency compared to this baseline, our changes are limited to the encoder and leave the hyperprior and analysis networks unchanged.} 

    \label{icml-historical} 

    \end{center} 

    \vskip -0.2in 

\end{figure*} 

\section{Proposed method}

Resource-constrained compute environments, such as mobile phones and wearables, increasingly use point clouds to represent three-dimensional objects. Compute on these devices is one of the most significant performance constraints. These use cases require point clouds to be sent across a network by devices with significantly fewer compute resources than where they are to be decoded. By implication, reducing parameter counts or reducing in total operations has broad implications for these use cases’ feasibility.

We propose a more efficient learned PCC encoder architecture, which we call DeepCompress. This architecture combines a better choice of activation and a more efficient convolutional block. In direct comparison with the current state of the art for operation-efficient learned point cloud compression (e.g., \citet{quach2020improved}) using dense convolutions, DeepCompress has a dramatic reduction in parameter and operation count with a minimal loss in quality. In later sections, we show that DeepCompress reduces total model parameter count by 20\% and reduces total convolution operation count by 8\%.

Following \citet{quach2020improved}, \citet{wang2019learned}, and \citet{wang2020multiscale}, the first step in building our learned PCC architecture is learning an arithmetic encoder with a scale hyperprior as shown in Fig. 1. The training is performed with a rate-distortion function by using a Lagrangian loss: 

\begin{equation}
L = R + \lambda*D 
\end{equation}

\begin{figure}[ht] 

    \vskip 0.2in 

    \begin{center} 

    \centerline{\includegraphics[width=\columnwidth]{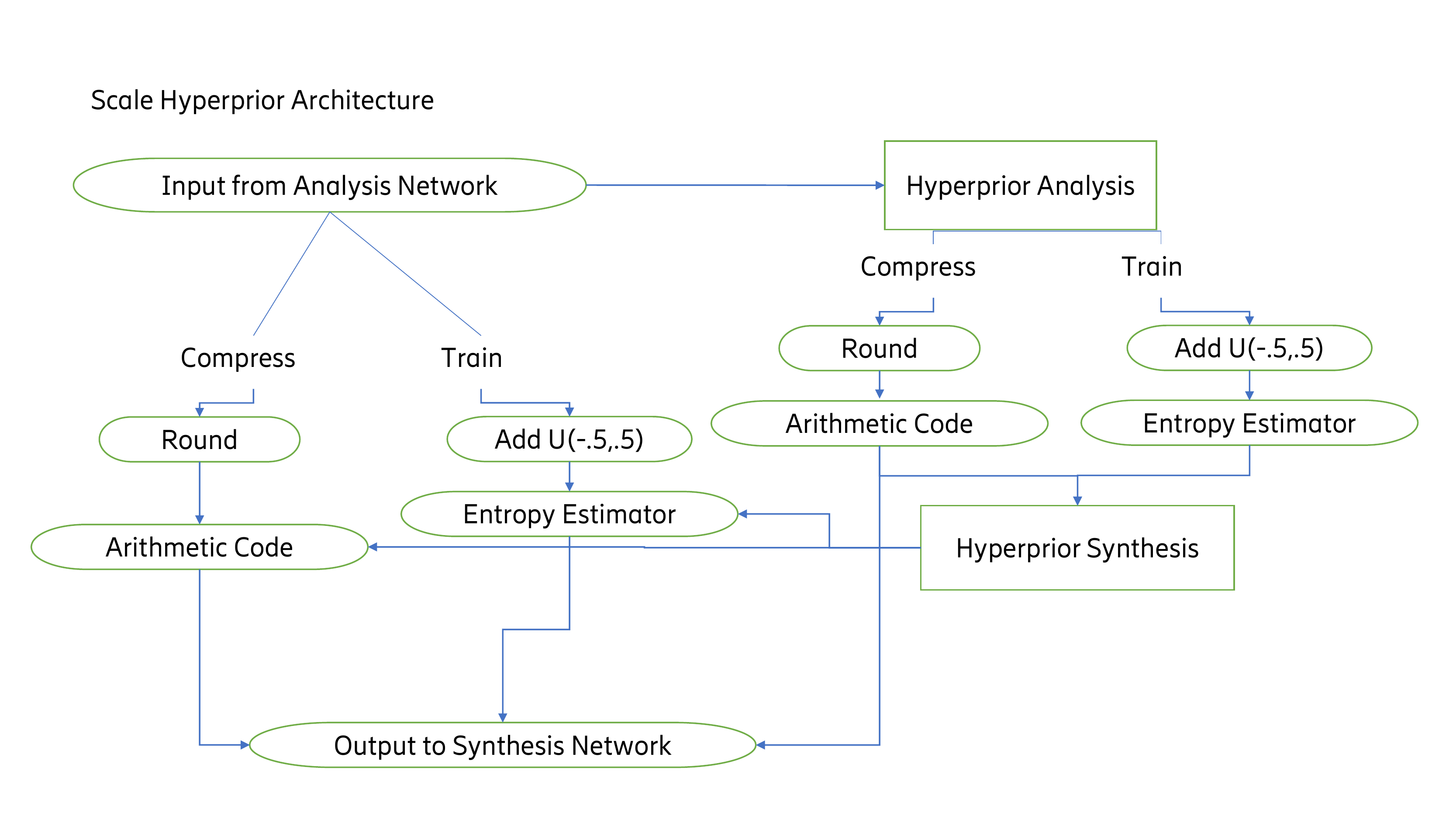}} 

    \caption{The scale hyperprior architecture encompassing our entropy model.} 

    \label{icml-historical} 

    \end{center} 

    \vskip -0.2in 

\end{figure}

\begin{figure}[ht] 

    \vskip 0.2in 

    \begin{center} 

    \centerline{\includegraphics[width=\columnwidth]{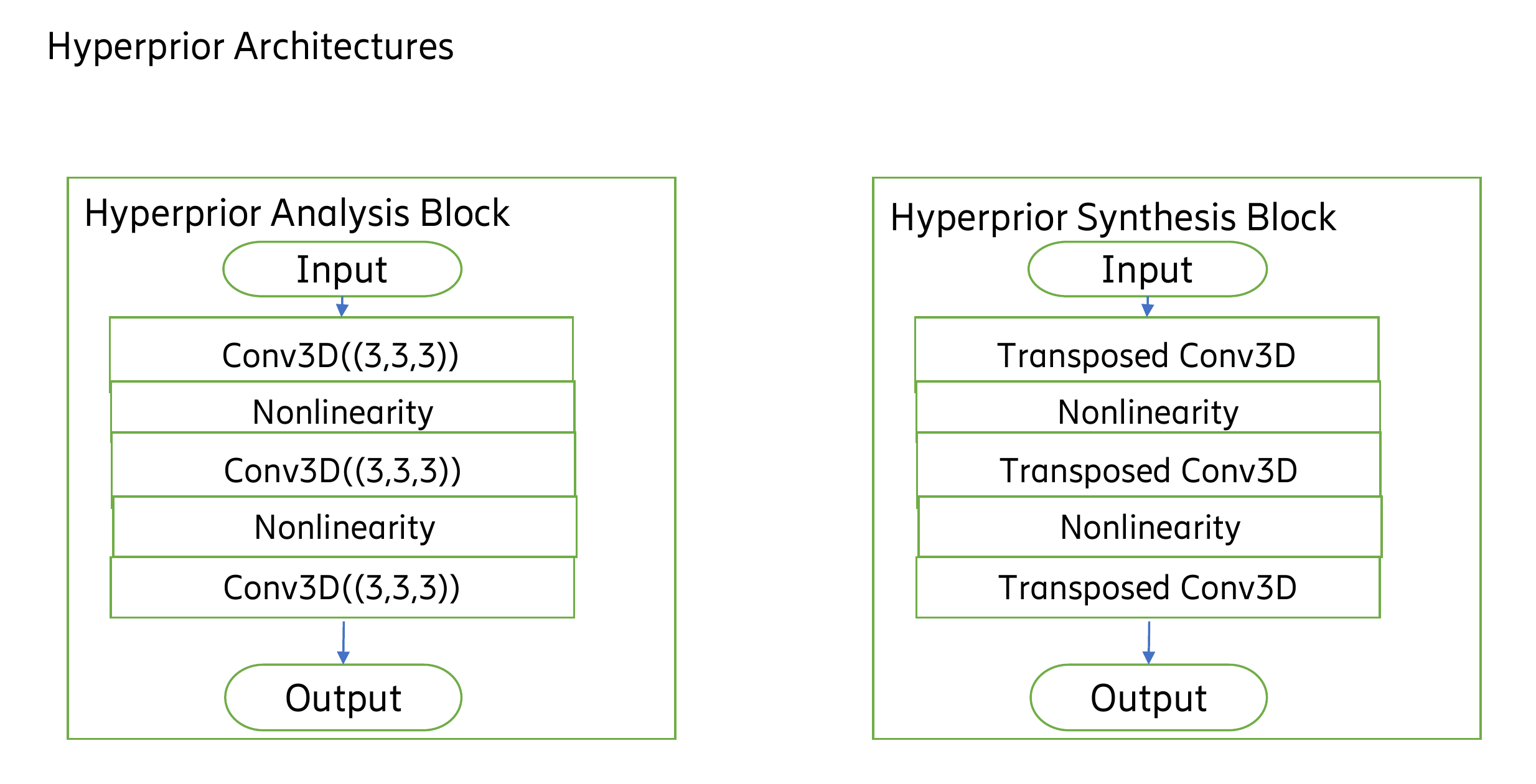}} 

    \caption{The standard convolutional architectures used in scale hyperprior models. Because we are exclusively concerned with the primary network architecture use ReLU nonlinearities across all of our models for these blocks.} 

    \label{icml-historical} 

    \end{center} 

    \vskip -0.2in 

\end{figure}

Which trades off a rate term ($R$), for a loss-term ($D$). This scale hyperprior network is illustrated in Figs. 3 and 4. 

Learned PCC architectures incorporate a learned scale hyperprior by setting the prior probability distribution used to produce the code in the primary channel to a standard zero-mean Gaussian distribution with a side-channel hyperprior code. This code then transmits information containing standard deviations for the primary channel, which in turn provides the probability distribution for the entropy estimation/coding of the latent variable in the primary channel. 

Expressed formally, our approach is as follows: 

\begin{equation} 
y = f_a(x), \tilde{y} = Q(y), \tilde{x} = f_s(\tilde{y}) 
\end{equation} 

Where $f_a$ and $f_s$ are the analysis and synthesis networks respectively. With the addition of a code, which is a side channel to carry information that directly sets a learned scale for the zero-mean Gaussian prior of $f_a$: 

\begin{equation} 
z = f_{ha}(y), \tilde{z} = Q(z), \sigma = f_{hs}(z) 
\end{equation}

 For our reconstruction loss, we consider the state-of-the-art for non-multi scale reconstruction losses proposed in \citet{quach2020improved}.

\begin{equation} 
\mathsf{focal\_loss}(x, \tilde{x}) = \alpha_t x_t \left(1 - \tilde{x_t}\right)^\gamma log(\tilde{x_t}) 
\end{equation}

Our loss is a focal loss with parameters $\alpha,\gamma$ which alleviates difficulties associated with predicting the highly sparse outputs necessary in point cloud reconstruction. Adopting the notation from \citet{quach2020improved}, we set $x_t = x, \alpha_t = \alpha, \tilde{x_t} = \tilde{x}$ if $x=1$ and 1 minus the right-hand side quantities if $x = 0$.

\begin{figure}[ht] 

    \vskip 0.2in 

    \begin{center} 

    \centerline{\includegraphics[width=\columnwidth]{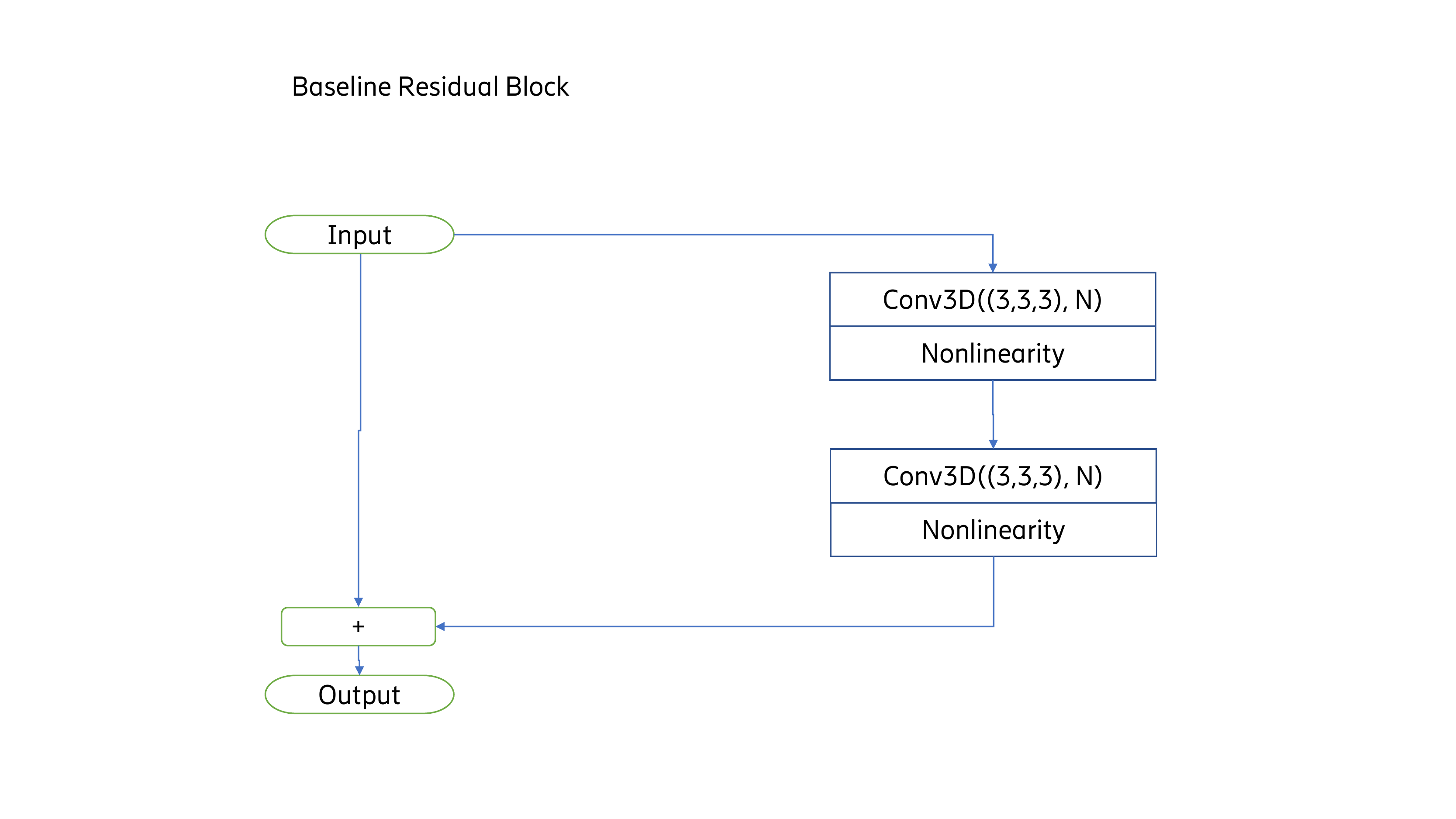}} 

    \caption{The residual block used in the baseline network. This is applied once per Analysis Block.} 

    \label{icml-historical} 

    \end{center} 

    \vskip -0.2in 

\end{figure}










\begin{figure}[ht] 

    \vskip 0.2in 

    \begin{center} 

    \centerline{\includegraphics[width=\columnwidth]{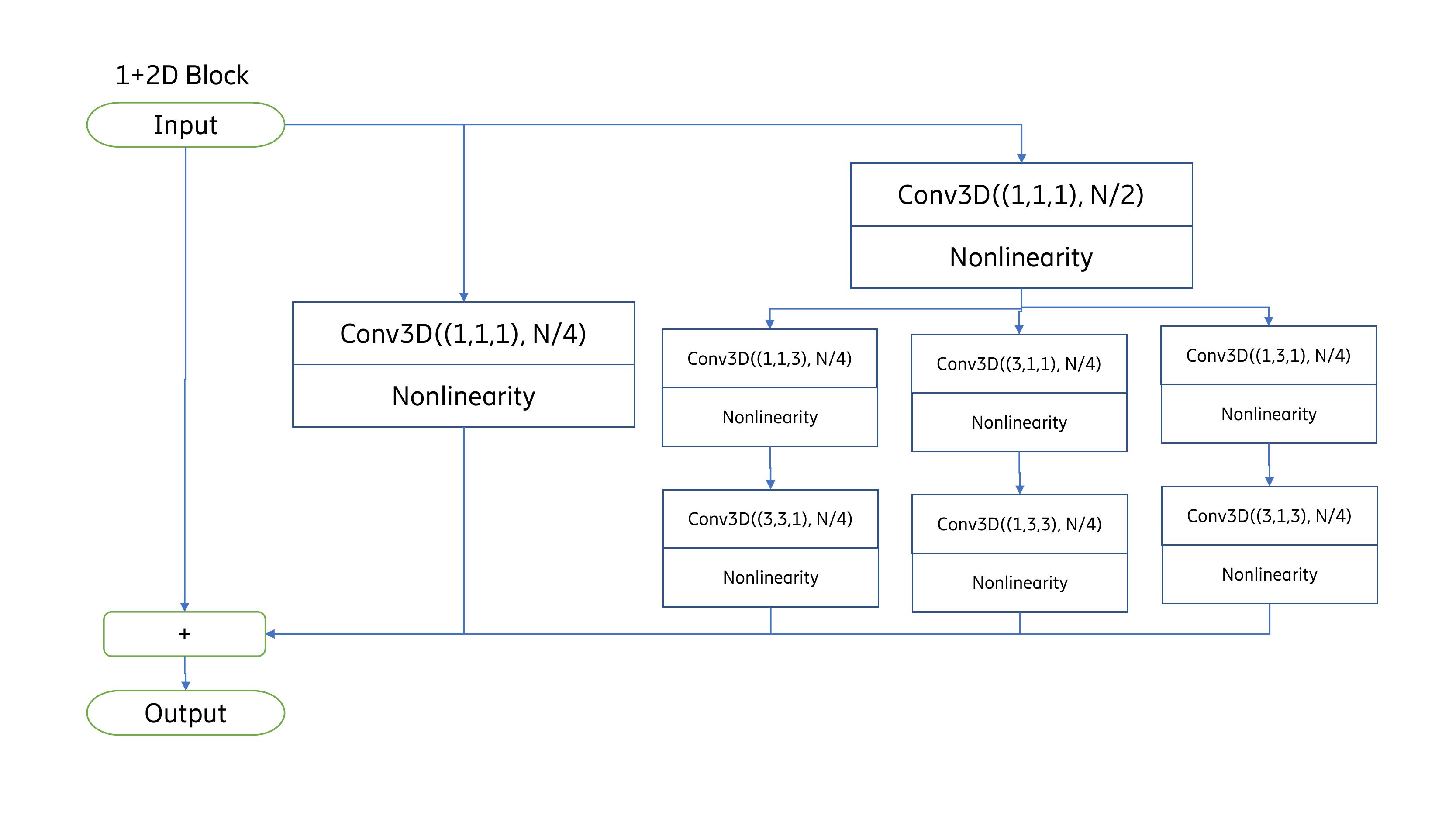}} 

    \caption{Our proposed block} 

    \label{icml-historical} 

    \end{center} 

    \vskip -0.2in 

\end{figure}

Building on InceptionV4, our proposed architecture shown in Fig. 5 combines a purely residual path — a 1x1x1 conv on a dedicated path — and distinct paths for factorized convolutions. In keeping with the standard practice in learned lossy compression and unlike Inceptionv4, we do not apply Batch Norm. This is because we find this destabilizes training.\footnote{In an experiment, we find that incorporating a BatchNorm layer made training fail to progress past intermediate loss values and can be taken (e.g., by \citet{balle2016endtoend}) to be a less powerful version of GDN in its effectiveness.}

Each of our Conv layers include a convolution using the parameters denoted and the as-written kernel size. The network’s default choice of non-linearity follows each Conv layer in our final result CGDN, which is described above. 

The proposed block in Fig. 6 is then utilized to replace the residual block of the analysis block in Fig. 1. In our proposed design, we only replace the third analysis block, but we also show that with only a mild reduction in performance we can replace the 2nd residual block as well. Fig. 4 illustrates the original baseline network that is used in \citet{quach2020improved}.

\subsection{Guiding design criteria} 

Learned PCC must incorporate several pieces of domain knowledge about natural object data when representing point clouds. Unlike \citet{xie2017aggregated} and similar architectures, we cannot favor a specific dimension when designing DeepCompress’ architecture. This is because we cannot discount that features will differ on properly oriented point cloud data in ways that are axis-dependent, such that it does not enable parameter-sharing as in \citet{kim2020triplanar}. Unlike in 2D compression, we also cannot assume \emph{a priori} the existence of a dominant dimension in three-dimensional depictions of objects or scenes. 

We argue that 1+2D convolutions are preferable to 3D when representational power in PCC is concerned. This is because: 

\begin{itemize} 

\item They are more parameter efficient given the same number of input/output channels 

\item They require fewer operations 

\item They more fully utilize their filters 

\item They encode the knowledge that point cloud data is a 2D surface in 3D space 

\end{itemize}

\subsection{Efficiency of the proposed model}

The total number of parameters in a 3D convolution for $x,y,z$ being the kernel size in each dimension, 

$C_i$ being the number of input channels, and $C_o$ the number of outputs is $xyzC_iC_o$ plus bias, if applicable, which takes $C_o$ parameters. Therefore, the minimal 3D kernel size of 3 which has been shown to work the best in PCC architectures that \citet{quach2020improved} takes $27C_iC_o$ parameters to represent. Compared to a sequential application of 1D and 2D convolution, which take only $3C_iC_o + 9C_o^2$ parameters if the 1D convolution is applied first. 

Assuming a cube of dimension $D$ as input in each channel, the number of operations necessary to perform a 3D convolution is proportional to $xyzC_iC_oD^3$. For kernel size 3, it is $27C_iC_oD^3$. A significant difference between 3D and 2D architectures is that it is considerably easier to incur a high operation cost by applying too many convolutions to large 3D volumes than large 2D images. This is demonstrated in the difference between our baseline shown in Fig. 5 and the Learned-PCGC \citep{wang2020multiscale} architecture shown in Fig. 6. This architecture uses significantly fewer parameters, but consumes more resources at run time as a result of these costly convolutions that are applied before downscaling.

Performing the same analysis for operational efficiency on our 1D + 2D convolution pairing, we see the number of operations is proportional to $xC_iC_oD^3 + yzC_o^2D^3$, which for kernel size 3 is $3C_iC_oD^3 + 9C_o^2D^3$. Because in our case we use an N/2 channel input and an N/4 channel output, our nonlinear 1+2D convolutions have in total $(\frac{3}{8} + \frac{9}{16})N^2D^3$ operations due to convolution to the $\frac{27}{8}N^2D^3$ necessary if they were full 3D convolutions with the same number of input and output channels and the same kernel size. 

We compare the costs between architectures in parameters and operations in terms of the dominant cost - convolutions - in Table 1.










\begin{table}[]
\caption{Comparison of Encoder Architectures by Conv Costs.}
\vskip 0.15in
\begin{center}
\begin{small}
\begin{sc}
\begin{tabular}{lll} 
\toprule
Architecture  & Operations & Parameters \\ 
\midrule
Baseline              & 1.118B               & 802k                 \\ 

Learned PCGC          & 5.233B               & 311k                 \\ 

Proposed              & 1.02B                & 610k                 \\ 

Proposed w/ 2nd block & 823M                 & 562k                 \\    
\bottomrule
\end{tabular} 
\end{sc}
\end{small}
\end{center}
\vskip -0.1in
\end{table} 

\subsection{Fuller filter utilization} 

A desirable property of convolutional neural networks is that they apply translational invariance to reduce parameter counts. They also pad the inputs to each convolution with zeros with the assumption that most convolutions will not be on the border of the region that is being filtered. This procedure of zero-padding causes problems in later layers of PCC architectures. The curse of dimensionality implies that higher-dimensional convolutions will have a vanishingly small fraction of the input volume that fully utilizes the representational power of the full convolution. 

To prove this mathematically, consider a kernel of size 3. For a D-dimensional convolution on an input of dimension size N, $\frac{N-2}{N}^D$ will be the fraction of the input that fully utilizes the parameter count. This is because the input volume is padded with zeroes that do not add additional information, but mask out parts of the filter. The absolute corners mask the largest fraction ($\frac{5}{9}$) of the parameters in each filter, however the ideal situation which we would like to achieve is to have an extremely small fraction of the input produce outputs which do not depend upon a fraction of the filter weights. 

For large N, this exponential reduction will be insignificant, but for N=8 and N=16, it implies that the usual parameter efficiency gained through translation invariance is reduced significantly for 3D convolutions in comparison to 2D and 1D.

\subsection{2D surface inductive bias} 

Our dataset is comprised of 2D surface information represented by occupancy maps in voxelized spatial regions. This means that not only is it sparse in nature, but that it does not include points that are not on the surface of the objects being compressed. We know that our input data will roughly correspond to orientations that are specific to how an object would naturally appear. As a result, we expect that much of the data will align roughly with one of the axes along which it is oriented, or that the lack of alignment can be compensated for by incorporation of perspectives from multiple such alignments. 

Similar reasoning that natural projections and orientations are good representations for 3D PCC is used when applying V-PCC to non-voxelized point cloud data. By this reasoning, our 1D convolution followed by a nonlinearity can be possibly seen as aggregating information in the dimension normal to our 2D convolution

\section{Experimental set-up} 

Similar to \citet{quach2019learning}, we train our model on a subset of the ModelNet40 dataset. We use the same preprocessing techniques to allow direct comparability and take the 200 (by total number of points) largest point clouds from the ModelNet40 dataset at a resolution of 512. We then eliminate all but the 4,000 cubes of size 64 with the most points. We consider these point clouds quantized at a resolution of 512 to create our overarching octrees, which terminate at input sizes of $64\times64\times64$ to produce the inputs to our network.  

Our experiment then iterates over rate-distortion lambda values ranging from $5 \times 10^{-5}$ to $3 \times 10^{-4}$ and then evaluates with additional MPEG-standardized G-PCC trisoup and octree as baselines alongside \citep{quach2019learning, quach2020improved} trained on the same data. The experiment uses Point Cloud Library and MPEG PCC dmetrics suite for the D1 and D2 metrics, Python 3.6.9 and TensorFlow 1.15.0 with the Adam optimizer with parameters (0.9, 0.999) \citep{kingma2015adam}. We use a separate learning rate and optimizer of $1\times10^{-4}, 1\times10^{-3}$ for the reconstruction and entropy bottleneck terms respectively.

We express the results from both of our experiments in terms of Chamfer’s Distance PSNR (D1 PSNR), and Point to Plane Distance PSNR (D2 PSNR) \citep{schwarz2018common}. Both methods rely on pairing each point in the reconstruction $b_i$ to the nearest point in the reference point cloud $a_j$ and vice-versa. The D1 distance simply returns the mean squared error (MSE) of the set of differences, whereas the D2 distance takes into account the nearest neighbors in the reference point cloud and projects along the normal direction before taking the MSE. To convert from a distance to a PSNR, the maximum possible distance between points is incorporated as $p$, which for all of our $1024 \times 1024 \times 1024$ reference point clouds is 1023.   

To compute these values, we use the MPEG PCC dmetric suite with normal provided by the nearest 12 neighbors as recommended in the MPEG CTC guidelines using the point cloud library (pcl) toolkit \citep{schwarz2018common}. 

Inherited from \citet{wang2019learned}, our thresholding method allows us to dynamically reduce the density of the reconstructed point cloud in practice. At test time, we likewise evaluate the performance of our model on the JPEG Pleno Voxelized Full Body (8iVFB v2) dataset \citep{d2019jpeg}. This dataset consists of four high resolution point clouds of humans in various poses and is the standard evaluation dataset for PCC. We compare our results to those reported in \citet{quach2020improved}, and use G-PCC as a reference against which to compare our performance gains in bit rate. 

\begin{figure}[ht] 

    \vskip 0.2in 

    \begin{center} 

    \centerline{\includegraphics[width=\columnwidth]{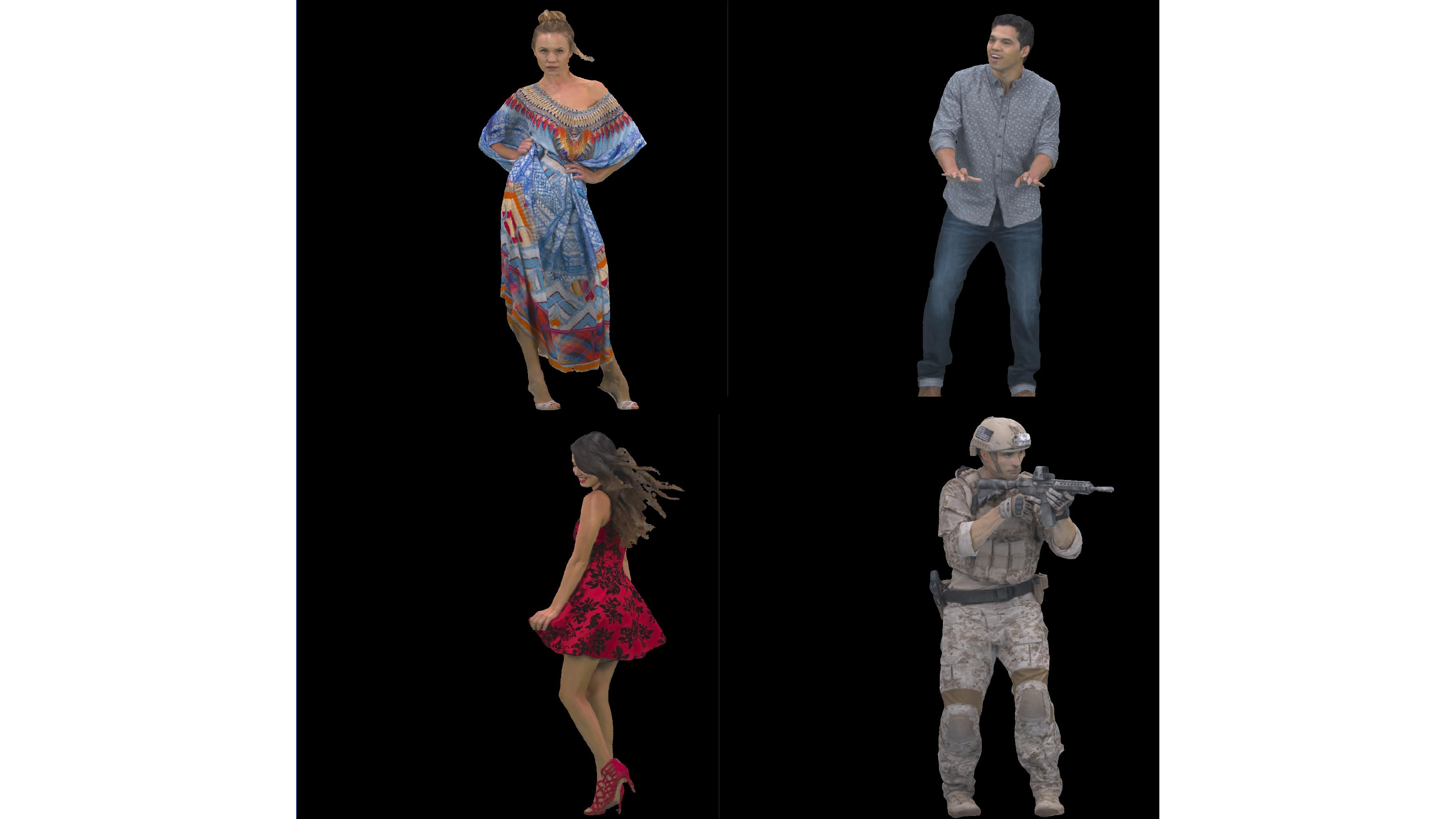}} 

    \caption{Pleno 8iVFB full body images with texture for detail (not compressed). 
    From left to right, top to bottom, they are: longdress\_vox10\_1300, loot\_vox10\_1200, 
    redandblack\_vox10\_1490, and soldier\_vox10\_0690} 

    \label{icml-historical} 

    \end{center} 

    \vskip -0.2in 

\end{figure}

We refer to these samples throughout the remainder of this paper by their given names. They are: loot\_vox10\_1200, longdress\_vox10\_1300, redandblack\_vox10\_1490, and soldier\_vox10\_0690.  

Although we train DeepCompress — our DNN model — on a subset of ModelNet40. Not only does this facilitate comparisons, training on ModelNet40 increases our model’s robustness and external validity to more complex point clouds. We find that data from ModelNet40 do not compress as well as those from the 8iVFB v2 dataset. This is likely because both Shapenet and 8iVFB v2 are considerably smoother in nature than the common household objects found in ModelNet40. Not only does this facilitate comparisons, training on ModelNet40 increases our model’s robustness and external validity to more complex point clouds. 

To train our model to output a highly optimized arithmetic encoder decoder pair and to ensure its stability, we train all of our models using TensorFlow-probability. This is a freely available library that automatically performs entropy estimation and prior-distribution learning.  

For both of our experiments mentioned in the below subsection, we employ the warm-start strategy advocated by \citep{quach2020improved}. We similarly set our loss’s $\alpha$ value to 0.75, and $\gamma$ equal to 2 in Equ. (4). We train our model until convergence using batch sizes of 32 at each lambda setting. We determine convergence by performing a validation step consisting of recording the mean loss over 10 mini batches on a subset of the ModelNet40 test set constructed in the same process as the training set. We determine convergence to have occurred after 3 consecutive validation steps occur with no reduction in mean loss. 

To encode our point cloud data, we first create an octree structure, which we store losslessly encoded alongside all $64\times64\times64$ blocks which we compress using our (hyper/)analysis network and learned arithmetic coder separately. Additionally, we transmit a single 32-bit number containing how many points are in each block per block to be used as a threshold. To decode we use our learned arithmetic decoder and (hyper/)synthesis networks to produce magnitudes which we threshold using the number of points for the input. Finally, we use the octree structure to reassemble all decompressed volumes within the larger point cloud. 

\subsection{Experiment 1: ReLU vs GDN vs CENIC GDN} 

Residual Linear Units (ReLU) are used in all other voxel-volume based neural network PCC architectures  of which we are aware \citep{wang2019learned, wang2020multiscale, quach2019learning, quach2020improved}. This is due to their incorporation in standard architectural building blocks such as Voxception Residual Network (VRN) \citet{brock2016generative} and ResNet \citet{he2015deep}. GDN has widely been used for image compression as an nonlinear layer and shown to outperform the ReLU layer \citep{balle2016density}. This improved performance obtained are that not only are ReLU units not invertible, which theoretically implies that they would inefficiently store information in each channel, but they are also poor at the task of transforming arbitrary data into per-channel Gaussianized input to the learned arithmetic encoder. GDN was originally proposed to solve this issue. 

The GDN activation is expressed mathematically as a parametric equation with variables $\alpha, \beta, \gamma, \epsilon$ as follows: 

\begin{equation} 
GDN(x) = \frac{x_i}{\left(\beta_i + \sum_j \gamma_{i,j}|x_j|^{\alpha_{i,j}}\right)^\epsilon} 
\end{equation}

Through a simple replacement of all ReLU activations in our encoder with GDN activations we are able to realize an improvement in performance that is similar to what is seen in more traditional learned image compression models, despite the acknowledged differences in the nature of our data. The originally proposed GDN activations which we tested fixed $\alpha$ to equal 2 and $\epsilon$ to equal .5 so as to normalize according to an $\ell^2$ norm. Furthermore, we can apply CENIC-GDN \citep{johnston2019computationally} activations which support a strict subset of the activations which can be represented by GDNs, and are significantly faster. This is because they fix $\alpha$ and $\epsilon$ to equal one, which converts the power operations to a simple absolute value. 

\begin{equation} 
CGDN(x) = \frac{x_i}{\beta_i + \sum_j \gamma_{i,j} |x_j|} 
\end{equation}

The impact on total parameter counts and total operations of switching from nonparametric ReLU activations to CGDN activations is extremely small. This is because they are only applied as activations after convolution has taken place and only require parameters dependent upon the number of output channels with a similar cost to a $1\times1\times1$ convolution.

In this experiment, we compare the performance of the compression efficiency by using ReLu and CGDN for a baseline residual block shown in Fig. 4 

\subsection{Experiment 2: Incorporation of 2+1D block}

To reduce the total number of parameters in our architecture by the maximum possible amount while maintaining functional parity in all other aspects, we isolate the last analysis block as shown in Fig. 1 with the maximal number of channels. We replace the residual component, which contains a single instance of the baseline residual block, with two copies of our proposed residual block as depicted in Fig. 6, and refer to this as the proposed method. When we apply our proposed blocks at the highest resolution, we significantly reduce the model’s performance. With only modest degradation in performance, we also show that the second-to-last residual block can also be replaced, and refer to this as the proposed method 2.

We do not change the architecture of the non-residual convolutions before or after our modifications as they are used to downsample the input. In the case of the final convolution, providing an arbitrary linear filter before quantization is performed and as such it is incompatible with our dimension-and-channel preserving nonlinear transform. We then train using GDN activations established in the prior experiment. In so doing, we reduce the total encoder (including hyperprior encoder) architecture's parameter count from 806,888 to 639,192, a 20\% reduction. We also reduce the number of operations due to convolutions significantly, yielding a 8\% reduction in computation. The magnitude of these differences is due to the fact that at each successive reduction in dimension we reduce total convolutional operator count by a factor of two. 

\section{Experimental Results}

\begin{figure}[ht] 

    \vskip 0.2in 

    \begin{center} 

    \centerline{\includegraphics[width=\columnwidth]{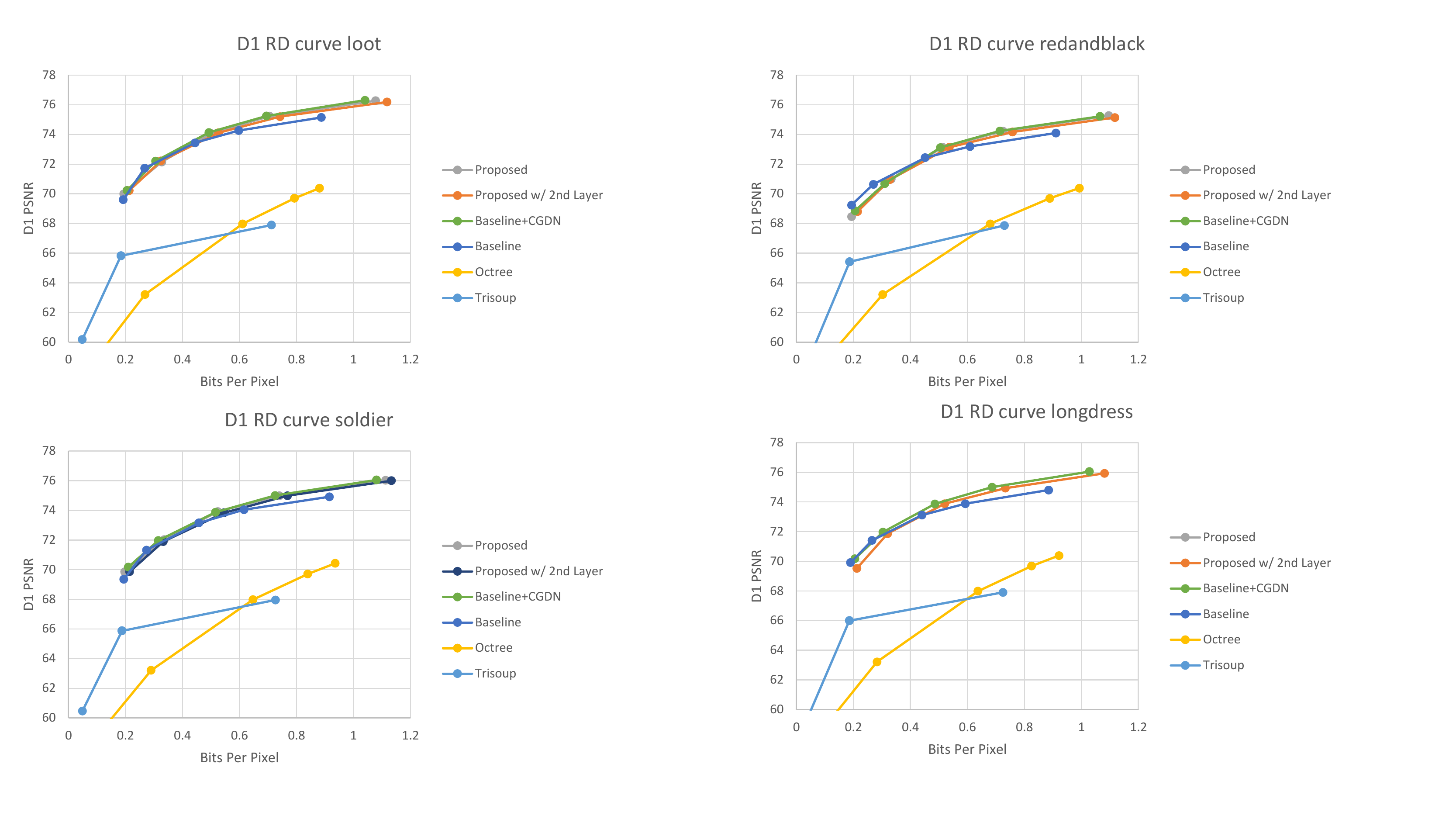}} 

    \caption{The plot of D1 PSNR vs Bits per point for each reference model.} 

    \label{icml-historical} 

    \end{center} 

    \vskip -0.2in 

\end{figure}

\begin{figure}[ht] 

    \vskip 0.2in 

    \begin{center} 

    \centerline{\includegraphics[width=\columnwidth]{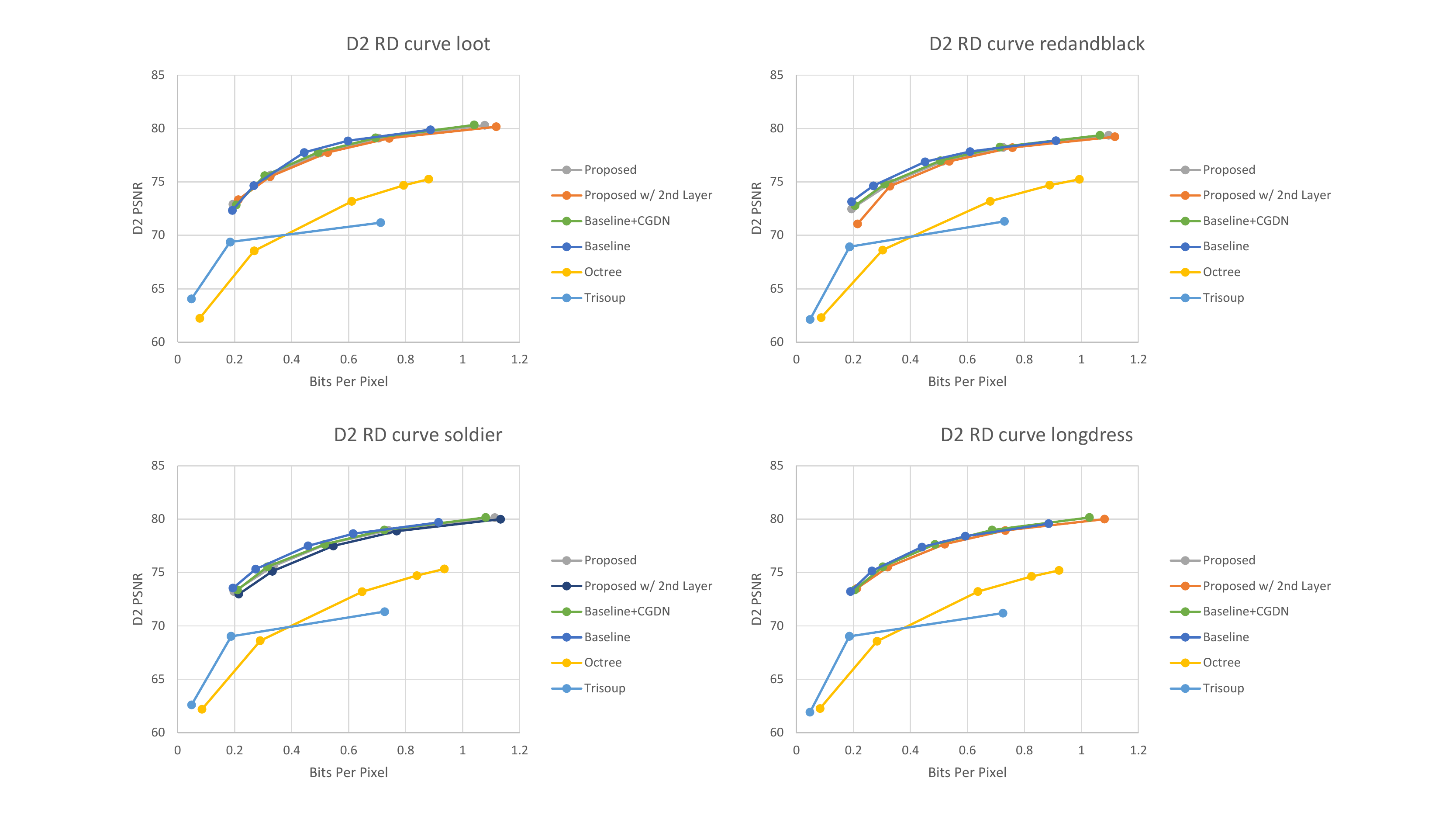}} 

    \caption{The plot of D2 PSNR vs Bits per point for each reference model.} 

    \label{icml-historical} 

    \end{center} 

    \vskip -0.2in 

\end{figure}  

\begin{table*}[t]
    \caption{(D1, D2) BD-rates relative to G-PCC.}
    \label{BD-rate comparison}
    \vskip 0.15in
    \begin{center}
    \begin{small}
    \begin{sc}
    \begin{tabular}{lcccc}
    \toprule
    Model & loot & longdress & soldier & redandblack \\
    \midrule
    Baseline+CGDN & (-76.22\%, -65.76\%) & (-76.86\%, -69.05\%) & (-76.51\%, -68.38\%) & (-72.09\%,	-66.46\%) \\
    Proposed      & (-76.22\%, -66.31\%) & (-76.74\%, -68.54\%) & (-76.51\%, -68.38\%) & (-72.09\%,	-66.46\%)  \\
    Proposed 2\footnote{Proposed w/ 2nd Block}    & (-75.12\%, -65.54\%) & (-73.69\%, -67.86\%) & (-74.67\%, -64.28\%) & (-71.49\%,	-60.00\%)  \\
    \bottomrule
    \end{tabular}
    \end{sc}
    \end{small}
    \end{center}
    \vskip -0.1in
    \end{table*}

As is evident from our plots of PSNR against bits per point for the PCC geometry, our incorporation of CGDN activations noticeably improves performance, see Fig. 7 for the Baseline+CGDN over the Baseline case. This is particularly with respect to the D1 metric. Additionally, there is no significant difference in quality reduction in terms of PSNR between our proposed architecture with 639,000 parameters and the baseline with 806,000, see Fig. 7 and Fig. 8 and Table 2 which compares our performance to G-PCC formatted to show (parenthesized to group) D1 and D2 BD rates for each reference model.

\section{Conclusion and Future Work} 

This paper contributes to a growing literature applying deep neural networks (DNNs) to learned compression of three-dimensional point cloud data. Because they flexibly represent 3D space, point clouds are an increasingly common in multiple fields. These include environmental understanding, object detection and tracking, and 3D reconstruction. Many of these use cases are require compute in resource constrained environments, such as mobile phones and wearable devices. This motivates our focus on computational efficiency.

Our model, DeepCompress, takes a principled approach to applying DNN to learned point cloud compression (PCC). Compared to existing techniques, our model’s architecture uses lightweight convolutional blocks. This approximation only slightly lowers the performance while dramatically increasing computational efficiency. DeepCompress reduces the total number of operations by 8\% and total model parameters by 20\%. We measure performance using the Bjontegard delta (BD) rate. Our architecture has a penalty of 0.02\% in Chamfer’s Distance (D1) and 0.32\% increased bit rate in Point to Plane Distance (D2) when compared to G-PCC, resulting in a trade-off between efficiency and bit rate.

Noting that multiscale understanding is necessary to scale convolutional PCC models to larger data, improve fidelity, and reduce operation costs \citep{wang2020multiscale}, a promising research direction is to build upon our architectural modifications to incorporate conditional information from multiple scales, potentially even embedding this process in the architecture itself by drawing inspiration from research works on image variational autoencoders \citep{zhou2018variational}. 

Furthermore, analysis of other factors that contribute to learned PCC performance is necessary to fully realize the promise of this technique. That includes data augmentation, proper dataset selection, and online or single-shot learning to make full use of the learned nature of these point cloud compression algorithms \citep{zou20202}. 

Another research direction is to consider device-centric optimization factors such as device energy efficiency and device-specific latency within an end-to-end PCC framework. 

\newpage
\bibliographystyle{abbrvnat}
\bibliography{example_paper.bib}  






\end{document}